\documentclass{article}

\usepackage{arxiv}

\usepackage[]{graphicx}
\usepackage{amsmath}
\usepackage{amssymb}
\usepackage{epsfig}
\usepackage{epstopdf}
\usepackage{subfigure}
\usepackage{algorithm}
\usepackage{algorithmic}
\usepackage{booktabs}
\usepackage{enumitem}
\usepackage{multirow}

\usepackage{tikzsymbols}

\title{Effect of Text Processing Steps on Twitter Sentiment Classification using Word Embedding}

\author{Manar D. Samad and Nalin Khounviengxay\\
Department of Computer Science\\
 Tennessee State University\\
 Nashville, TN, USA\\
\texttt{msamad@tnstate.edu} \\
\and
 \bf {Megan A. Witherow}\\
 Department of Electrical and Computer Engineering\\
  Old Dominion University\\
   Norfolk, VA, USA }

\begin{document}

\maketitle

%
%
%
%(e-mail: ndiawara@odu.edu)}

%\IEEEpeerreviewmaketitle

%to achieving the crucial first step that largely determines the final text classification accuracy.

\begin{abstract}

Processing of raw text is the crucial first step in text classification and sentiment analysis. However,  text processing steps are often performed using off-the-shelf routines and pre-built word dictionaries without optimizing for domain, application, and context. This paper investigates the effect of seven text processing scenarios on a particular text domain (Twitter) and application (sentiment classification). Skip gram-based word embeddings are developed to include Twitter colloquial words, emojis, and hashtag keywords that are often removed for being unavailable in conventional literature corpora. Our experiments reveal negative effects on sentiment classification of two common text processing steps: 1) stop word removal and 2) averaging of word vectors to represent individual tweets. New effective steps for 1) including  non-ASCII emoji characters, 2) measuring word importance from word embedding,  3) aggregating word vectors into a tweet embedding, and 4) developing linearly separable feature space have been proposed to optimize the sentiment classification pipeline.  The best combination of text processing steps yields the highest average area under the curve (AUC) of 88.4 ($\pm$0.4) in classifying 14,640 tweets with three sentiment labels. Word selection from context-driven word embedding reveals that only the ten most important words in Tweets cumulatively yield over 98\% of the maximum  accuracy. Results demonstrate a means for data-driven selection of important words in tweet classification as opposed to using pre-built word dictionaries. The proposed tweet embedding is robust to and alleviates the need for several text processing steps.  

\end{abstract}


\section{Introduction}

The last decade has seen an unprecedented surge in the growth and proliferation of online social media platforms such as Twitter, Facebook, and Tumblr. For individuals, social media is a uniquely powerful platform for communication, self-expression, microblogging, social networking, image and video sharing, and opinion sharing through online word-of-mouth. As of January 2020, the world has more than 3.8 billion active social media users~\citep{KepiosPte.Ltd.2020} generating prodigious volumes of public opinion data. For businesses and organizations, social media provides access to consumer bases for targeted advertising and public opinion mining on products and services to inform strategic business plans. An important task of public opinion mining is sentiment analysis (SA) to disambiguate sentiment from data shared on social media platforms. SA performs sentiment classification to stratify social media posts based upon the polarity (positive, negative, or neutral) of the expressed opinion in the textual data~\citep{Kharde2016, WangX2019}. 

A mass-producer of textual data is the popular microblogging platform Twitter, which reports 166 million average daily active users in 2020 Q1~\citep {TwitterInc.2020}. Twitter's defining characteristic is its character-limited microblog posts, called 'tweets'. Tweets have been limited to 140 Unicode characters until 2017 when the limit has doubled to 280 characters~\citep{Boot2019}. Due to the character limitation, Twitter users prioritize efficiency when composing tweets by employing informal language, abbreviations, unconventional word spellings, symbols and numerals, as well as omitting parts-of-speech (POS) to convey the desired message~\citep{Boot2019}. Twitter's condensed format and associated usage characteristics make it a challenging domain for SA unlike mining standard literature and documents. This paper investigates the efficacy of common text processing and classification steps on Twitter-specific SA. We hypothesize that efficient use of word embedding with data-driven optimization can be robust to the selection of standard text processing routines in Twitter sentiment classification.  

\subsection {Background review}

The two primary approaches to SA are: 1) lexicon-based and 2) machine learning (ML) based methods~\citep{MARTINEZ-CAMARA2014, kolchyna2015twitter}. Lexicon-based methods use a sentiment dictionary (i.e. lexicon) that contains words labeled with their polarity to determine the overall sentiment of the social media post~\citep{kolchyna2015twitter}. Some popular lexicons include WordNet~\citep{PrincetonUniversity2010}, SentiWordNet~\citep{Esuli2013}, and opinion lexicon~\citep{Liu2017}. Since lexicon-based methods depend on the words used in the social media post being present in the sentiment dictionary, these methods suffer from low recall~\citep{MARTINEZ-CAMARA2014}. Two words of different polarity (good and bad) can appear in the same tweet that complicates SA in a particular context. Furthermore, the noisy nature of tweets and ambiguous usage of words across contexts makes the automatic generation of lexicons from Twitter data unreliable ~\citep{kolchyna2015twitter}. On the other hand, the ML approach uses supervised optimization to learn a function that performs the classification task based upon labeled training examples. However, the primary limitation of ML methods is the availability of labeled social media posts for training SA models.

%Notably, today's innovation in artificial intelligence (AI) and ML has been revolutionized by the rise of Big Data generated through mass worldwide daily activity on the Internet. Over 2.5 exabytes of data~\citep{DOMO2018} are created globally each day by more than 4.5 billion Internet users~\citep{KepiosPte.Ltd.2020}. 

SA of social media posts, especially from character-limited tweets, remains a nontrivial task. For example, sentiment dictionaries used in lexicon-based methods do not fully capture new domain and context specific terminologies~\citep{Kharde2016}. Abbreviated expressions such as 'b4', 'plz', 'sry', and 'gr8' are common in Tweets. The informal linguistic style and variety of irregular abbreviations prompted by the character-limited Twitter environment exacerbate the limitations of the lexicon-based approach. ML-based methods offer a favorable alternative to lexicon-based methods for Twitter SA. However, ML models require all text to be converted to representative numeric data. This text-to-numeric data conversion is not straightforward and requires interdisciplinary efforts involving linguistics.

Early ML approaches to Twitter SA rely on raw word representations called n-grams as the primary input features~\citep{MARTINEZ-CAMARA2014}. An n-gram is a sequence of n consecutive words such that single words and pairs of words are referred to as unigrams and bigrams, respectively. A tweet is represented in a feature vector of length equal to the number of n-grams such that each n-gram has its unique position in the vector. The vector positions contain frequency of corresponding n-grams or simply hold  '1' or '0' depending on the presence or absence of the n-gram in the tweet, respectively. This construction is heavily reliant on substantial text processing and careful feature selection steps to optimize the classifier model performance. For example, Jianqiang et al. have extensively studied the effect of six well known text processing steps on Twitter sentiment classification using n-gram representations~\citep{Jianqiang2017} as summarized in Table~\ref{tab:01}. The reported effects are mixed across n-gram models, classification types (binary versus three-way) and classification algorithms. Many studies routinely employ standard text processing steps, including stemming~\citep{Singh2016}, lemmatization~\citep{Dereza2018}, removal of stop words~\citep{Ahamed2020} and non-ASCII characters ~\citep{Jianqiang2018} assuming that these steps are essential for all text analysis applications.  In contrast, both Bao et al.~\citep{Bao2014} and Kolchyna et al.~\citep{kolchyna2015twitter} report that stemming and lemmatization reduce Twitter sentiment classification accuracy. Zhao et al. report no effect of removing numbers, URL, and stop words on sentiment classification~\citep{Zhao2015}. 

Recently, ML-derived contextual vector representations of text (text embedding) have created a new paradigm for ML-based SA research~\citep{Tang2016}. Several highly efficient text-to-numeric word embedding algorithms have been developed by field experts and industrial giants, including skip gram~\citep{Balikas2016}, GloVe~\citep{Pennington2014}, and fastText of facebook~\citep{fastText}. For example, the popular word2vec routine, developed by Mikolov et al.~\citep{Mikolov2013EfficientEO, church2017}, is a highly optimized Python package that implements the skip gram model for obtaining a contextually meaningful word embedding that enables subsequent text analysis. In contrast to word embeddings trained for a particular application, there are publicly-available word embeddings pre-trained from large corpora, such as the GloVe database ~\citep{PenningtonData2014}. However, such pre-trained word embeddings may not optimally capture the underlying context (sentiment towards a subject matter) and the domain (tweets)~\citep{yu-etal-2017-refining, LiuX2015}. While training word embeddings directly from the domain text is important, it requires careful selection of text processing steps.

To the best of our knowledge, the effects of different text processing steps have not been well studied in word embedding-based models for Twitter SA. We hypothesize that the importance of these text processing steps depends on the text domain (Tweets) and the contextual (sentiment analysis) representation of words in the feature vector obtained via word embedding. A robust word embedding is expected to be less sensitive to the selection of text processing steps and classifier models. 

%%%%
\begin{table}[t]
\centering
\caption{Effects of text processing steps on sentiment classification performance reported in~\citep{Jianqiang2017} using Bag-of-Words (BoW) n-gram features.}
\begin{tabular} {ll} \toprule
%\hline
Text processing & Effects on sentiment classification\\  
steps& classification performance \\
\midrule
Removing URL	& Barely affects the performance \\ \midrule
\multirow{ 4}{*}{Removing stop words} &	Useful in general. No effects on 3-way\\
& classification using n-gram features. \\
& positive effect on binary   \\
& classification with random forest. \\ \midrule
Removing random words & Severely affects the performance \\ \midrule
Removing numbers &	Useful with mixed findings \\ \midrule
\multirow{ 3}{*}{Reverting repetition letter} &	Useful except in binary classification \\
& using Support vector machine. \\
&No effects on 3-way classification. \\ \midrule
Expanding acronym &	Positive effects in general\\\midrule
\vspace{-30pt}
		\end{tabular}
	\label{tab:01}
\end{table}
%%%%%%%%%%%%%%%%%
A common text processing step is stop word removal using off-the-shelf stop word dictionaries. For example, the several decades old Van stop list~\citep{vanStop} is used in recent Twitter SA literature~\citep{Jianqiang2018}. In contrast, a pre-compiled list of stop words has shown negative effect on Twitter sentiment classification~\citep{Saif2014}. We hypothesize that the selection of stop words can be domain-specific to avoid discarding potentially informative keywords. Random removal of words has been reported to significantly deteriorate text classification performance~\citep{Jianqiang2017}. We propose that, prior to removing words, the importance of individual words can be learned from context-derived word embeddings. Thus, word selection or removal can be optimally made based on such data-driven decision instead of using a default dictionary of stop words.  

Furthermore, there are many non-dictionary words, symbols, and short and colloquial language expressions on social media that can complicate SA, and hence, are typically discarded from analysis. For example, 'wow', 'lol', 'omg', emoticons, and many other non-dictionary expressions frequently appear in tweets. The treatment of these non-dictionary but potentially informative words requires special attention. Previous studies have used emoticon dictionary~\citep{DCNN-34} to replace emoticons (composed of ASCII characters, such as :-) represents a happy expression) with their textual names (happy). Kolchyna et al. manually construct a lexicon of emoticons and social media abbreviations with associated polarities~\citep{kolchyna2015twitter}. 

The successors to emoticons, Unicode ideograms called emoji, such as \dSmiley, are used by Twitter users as multimodal markers to convey affect, gestures, or repetition, and/or to function semantically in a sentence by replacing parts of speech~\citep{Naaman2017, bai2019}. However, since text analysis is commonly performed only on ASCII characters, Unicode emojis are often filtered out. In Redmond et al., emojis have been shown to outperform emoticons in sentiment classification~\citep{redmond2017}. Manual~\citep{novak2015} and automatic~\citep{kimura2017} methods have been proposed to build emoji lexicons by labeling emojis with their corresponding sentiment. However, emoji lexicons are limited by the corpora without learning their contextual representations in word embeddings. We postulate that emojis may be directly represented using their individual UTF-8 encodings for learning the word embedding. In general, the use of off-the-shelf dictionaries or lexicons, including those used for stop word removal, emoji mapping, and obtaining pre-trained word embeddings may not provide flexibility for custom learning and domain-specific SA optimization. 

Last but not least, the strategy for aggregating multiple word vectors from word embeddings into a constant length  tweet vector is an open research problem. In machine learning, the ordering of constituent words in the tweet vector is important and informative. A simple concatenation of word vectors with zero padding~\citep{Jianqiang2018} or averaging of word vectors~\citep{Sarkar2016} is used in practice without identifying the best strategy. The effect of different word vector aggregation strategies needs to be evaluated on Twitter SA.

\subsection{Contributions}

Given the limitations in the current literature, this paper aims to investigate the effects of text processing and classification steps on Twitter SA. The contributions of this paper are as follows.
\begin{itemize} [leftmargin=0.15in]
\item Stop words and other non-ASCII or Unicode characters (Table~\ref{tab:02}) are routinely removed from corpora to emphasize more informative words. This exclusion may inadvertently leave out elements that can be informative for SA. More importantly, aggressive removal of words and characters can substantially reduce a character-limited tweet to only several or even no words. This substantial loss of data can adversely affect the performance of machine learning models. We hypothesize that retaining and properly encoding conventionally overlooked words and non-ASCII emojis can improve sentiment classification performance. 
\item Documents are often converted to feature vectors using an off-the-shelf word-embedding dictionary. These word-embeddings are developed using large corpora that may not capture atypical and domain-specific expressions of social media. Instead of using a global word-embedding dictionary, we propose to develop and analyze the word-embedding directly from the tweet data set to capture domain-specific contexts and trending keywords.   
\item For tweet classification purposes, machine learning models require all tweets to be represented in equal length vectors. However, each tweet constitutes varying numbers of words. Therefore, the challenge is to represent all tweets in an equal length feature vector by optimally aggregating the constituent word vectors. We compare several alternative methods of aggregating word vectors into a constant length tweet vector to identify the best strategy. 
\item We investigate if word embeddings can provide information about the word importance without the supervision of classification labels. We propose a data-driven method for estimating word importance to aid context informed selection or removal of words in Twitter SA.
\end{itemize}

The remainder of this paper is organized as follows. Section II describes the methodology towards our investigative study of text processing and classification steps in Twitter SA. Section 3 reports our results on the effect of several combinations of text processing steps, the dimensionality of the word vectors and word importance, and different tweet embedding methods. Section IV discusses the key findings and limitations of this paper. Section V provides concluding remarks.

\section {Methods}

Our proposed models are evaluated using a public data set of tweets on customer experience with several U.S. airlines~\citep{kaggle}. The data set is comprised of tweets from February 2015 when tweets are restricted to challenging 140-character limit. Each tweet is labeled by one of three sentiments: 1) positive, 2) negative, and 3) neutral for three-class classification of customer sentiments. The proposed processing and classification steps of tweets are discussed below.

\begin{table}[t]
\centering
\vspace{10pt}
\caption{Four processing steps for text analysis: 1) removing undesired portions of text by matching regular expression patterns, 2) stop words, 3) removing single or isolated characters using regular expression matching, 4) representing each emojis by its Unicode transformation format (UTF-8)}
		\begin{tabular} {ccl} \toprule
Step &Pattern& Description\\ 
\midrule
1 &  "htt$\backslash$S+$\backslash$ w*",   & Sub-string starting with (htt, ww) \\
   & "ww$\backslash$ S+$\backslash$w*"& followed by other characters. \\
   &&Examples: http, https, www, www.\\
   &"['  ’  .]"& Remove 'apostrophe', 'period'  \\
   &["?", “!”] & Isolate '?' and '!' using a space \\
    && in between the word and character.   \\
& "\&$\backslash$w+", "@$\backslash$w+" &Remove entire words starting \\
  &&     with '\&' and '@'. \\
& "[0-9]$\backslash$w+",  "[0-9]+"& Remove numeric strings \\
         &"$\backslash$w[0-9]+",&starting with a number or a letter \\
         &&examples: '8Goal', 'Go8al', '4567'.\\
         &$\-$, $\slash$, =, $\hat{}$, $\}$, :, $\{$, $\|$ &  List of non-alphabetic characters\\
         & +, $\ast$, " , $\sim$, ], [, $\hat{a}$, $\hat{e}$, $\tilde{a}$ & to be removed using regular\\
         & -, \#, …, $\)$, ;, \acute{a}, \acute{e}, \_ , $\($, ’& expressions.\\
         \midrule
  2&  Sample stop words to &['it', 'its', 'this', 'that',  \\
  & be removed &'are', 'was', 'were', \\
  & from corpus&'for', 'by', 'so', 'on', 'have', 'has',  \\
  && 'had', 'an', 'out', 'with', 'as', 'in'\\
  && 'be', 'been', 'to', 'the', 'and', 'of', \\
  && 'am', 'is', 'at', 'by', 'into']\\
  \midrule
  3 & tk =  &Remove single characters \\
  &[x for x in tk if len(x)$>$ 1] &from tokenized list of words (tk)\\
  \midrule
  4& Emojis & UTF-8 encoding of Emoji\\
  && \dSmiley is encoded and replaced by \\
  && (b'\textbackslash xf0\textbackslash x9f\textbackslash x98\textbackslash x8a')\\
  \midrule
                		\end{tabular}
	\label{tab:02}
	\vspace{-10pt}
\end{table}
\subsection {Text processing}
The text processing and cleaning steps are often arbitrarily chosen in the literature without systematic investigation of their effects on model performance. We study the effect of seven combinations of four text processing steps (Table~\ref{tab:01}) prior to obtaining the word embedding. First, regular expressions are used to remove undesired characters and words from tweets. Examples include hyperlinks, webpages, words starting with '@' and '\&', numeric values, and special characters.  Second, we create a list of stop words including auxiliary verbs, prepositions, and pronouns that are known to bear little or no semantic information. Note that this is an experimental list to study the effect of excluding stop words on tweet classification. Third, we exclude single characters from tweets. Fourth, we use the Unicode transformation format (UTF) of individual emojis in place of the Unicode character. 

\begin{figure*}[t]
%\normalsize
\centering
\includegraphics[trim=0cm 0cm 0cm 0cm, clip=true, totalheight=0.3\textheight, angle=0 ] {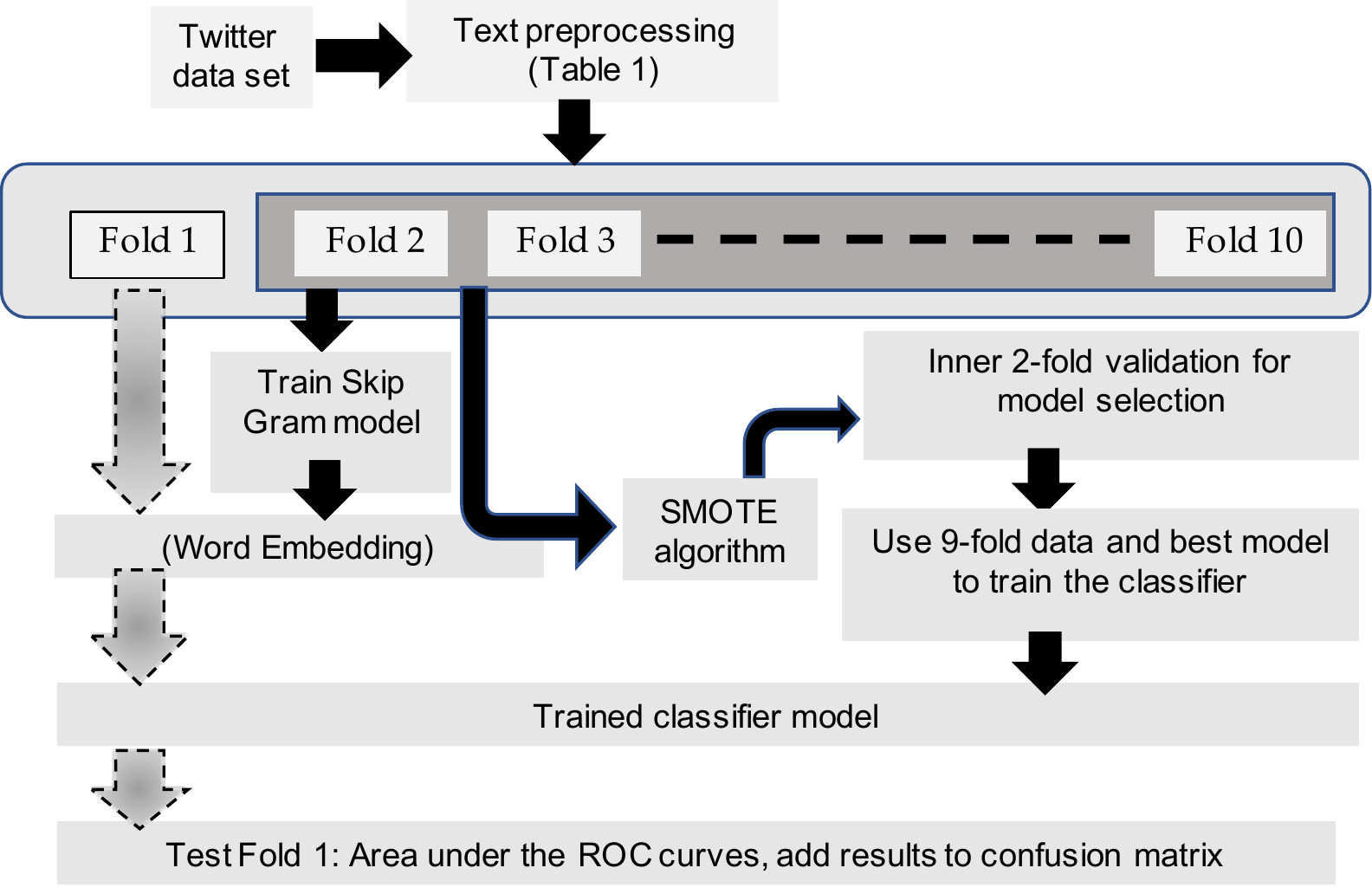}
\caption{Classifier model training, validation, and testing steps for one of the ten folds in 10$\times$2 nested cross-validation.}
%\vspace{-20pt}
\label{fig:01}
\end{figure*}

\subsection{Word embedding}

The first step of text classification is to encode semantic and contextual properties each word into a numeric vector. We use the skip gram model, developed within the highly optimized $word2vec$ package, to produce Twitter context-specific word embeddings. The word-embedding is a dictionary of numeric vectors for individual words in the corpus. The skip gram model is a neural network with a single hidden layer that uses a word vector in the input layer to predict a set of neighboring words in the output layer. This prediction learns the vector representation of the entire corpus vocabulary in the hidden layer as the word embedding. Given a vocabulary of N unique words in a Twitter corpus, each word can be represented in a N-dimensional one-hot encoded vector.  The one-hot encoded vector has a designated position for each word that contains '1' to represent the presence of the word and '0' otherwise. An autoencoder maps each N-dimensional input vector $X\in\Re^N$ (representing a single word) to a lower K-dimensional encoded space that can be used to reproduce the N-dimensional target output vector $Y\in\Re^N$ (representing one-hot encoded presence of multiple neighboring words in tweet). The encoding in the hidden layer, $Z\in \Re^{K}$, can be obtained by learning the word embedding matrix $W \in \Re^{KXN}$ as below,

\begin{equation}
Z = W X + Q1,
\end{equation}
%%%%%
such that Z can be mapped back to the known target Y vector as below. 
%%%%%%%%%
\begin{eqnarray}
Y &=& W^TZ + Q2, \nonumber  \\
Y &=& W^T (WX + Q1) + Q2
\end{eqnarray}

Here, Q1 and Q2 are bias vectors of K and N dimensions, respectively. Thus, the word embedding matrix $W^T$ has a K-dimensional vector representation for each of N words corresponding to N rows.

We do not use an off-the-shelf pre-trained word-embedding or a global corpus to obtain the word embedding. Instead, the proposed Twitter data set is used to create the word-embedding at each training cycle to capture the classification relevant context (airline sentiment) in tweets. This approach is also useful for text domains that may not have large training corpus or pre-built dictionary. For instance, predicting patient condition from physician’s generated text reports deals with many rare and domain-specific clinical terminologies. To strictly separate the training and test data, only the training data set (90\%) is used to yield the word embedding from the skip gram model. This word-embedding is used to extract feature vectors for both training (90\%) and test (10\%) data such that the skip gram model never sees the test data fold.  We further investigate the effect of skip gram window size and word vector dimensionality on sentiment classification accuracy.

%%%%%%%%%%%%%%%%%

\begin{algorithm}
\vspace{5pt}
\caption{Measuring word importance for word selection}
\begin{algorithmic}
\STATE Input: Word matrix, W = $[x]_{mx1}$, $x\in\Re^D$: a word vector
\STATE Output: 
\begin{itemize}
\item Set of m L2-norms, SL2 = $\{a\}^m_{i=1}$
\item Set of m variances, Svar = $\{b\}^m_{i=1}$
\item Sorted SL2 and Svar in descending order as Ssort
\item Select top k word in matrix, $W_k = [x]_{kx1}$, (k$\leq$m).
\end{itemize}

\STATE m $\leftarrow$ length (W)

\FOR {t =1 $\rightarrow$ m}
\STATE x  = W[t]  and then a = $\bf{xx^{T}}$
\STATE b = variance (x)
\STATE SL2 [t] = a and then Svar [t] = b
\ENDFOR
\STATE Repeat below separately for
\STATE Svar and SL2 in place of S
\FOR {r = 1 $\rightarrow$ m}
\STATE index [r]  = argmax (S)
\STATE maxValue = S [argmax (S)]
\STATE S$\leftarrow$ remove maxValue from S
\ENDFOR
\STATE Wk $\leftarrow$ W [index [1:k]]
\end{algorithmic}
\end{algorithm}

\subsection{Tweet-embedding and word importance}

Each tweet of variable word-length is required to be represented in a Tweet vector of constant length. The baseline approach is to average (Eq. 4) across n-dimensional m word vectors to yield an n-dimensional tweet vector.  We compare the performance of this baseline approach with three other methods of aggregation: 1) sum of vectors (Eq. 3), 2) weighted average of vectors (Eq. 5), and 3) selecting and adding the $k$ most important word vectors based on word importance. Let $m$ number of words in a tweet form a tweet matrix $[w_{ij}]\in\Re^{m\times D}$ where each row is a word vector.  We can obtain the tweet vector as below. 
%%%%%%%%
\begin{eqnarray}
Tw &=& \sum_{i=1}^{i=m} [wij], \\
Tw_{avg} &=& \frac{Tw}{m},\\
Tw_{weighted}&=& \frac{ \sum_{i=1}^{i=m} \frac{1}{1+ \sigma^2_j} [wij]}{\sum \frac{1}{\sigma^2_j}}
\end{eqnarray}
%%%%%%%%%%
where, $Tw\in\Re^D$ is a D-dimensional tweet vector and $\sigma^2_j$ represents the column variance of the $[w_{ij}]$ word matrix.
%%%%%%%
Next, we sort the words in descending order of importance to subsequently select the $k$ most important words. The cumulative contribution of the top $k$ most important words is investigated in a sentiment classification task. The word importance is measured using two metrics in an unsupervised way, i.e., being unaware of the relationship between the tweet and its sentiment class. The two metrics, the variance and L2 norm of individual word vector, are used to determine the relative importance of individual words.  Algorithm 1 shows the computational steps for selecting $k$ most important words.

\subsection {Model training and validation}

We use a 10$\times$2 fold nested cross-validation scheme~\citep{Varma2006} where the data samples are split 90\%-10\% (training-testing) in the 10-fold outer cross-validation as shown in Figure~\ref{fig:01}. The inner 2-fold cross-validation takes the 90\% data (outer nine folds) and splits it 50\%-50\% for model training and validation, respectively. The inner cross-validation yields the best hyperparameter values for the classifier model. Using the best hyperparameter values and the 90\% data, we train the model and then test on the left-out test data fold (10\%) in the outer cross-validation. This train-validate-test process is repeated for all ten data folds to obtain ten area under the receiver operating characteristic curve (AUC) values for classification. We use three one-versus-all classification models to perform the proposed three-class sentiment classification. For each sentiment class, the resultant AUC is obtained by averaging across ten AUC values corresponding to ten test data folds.

\begin{figure*}[t]
%\normalsize
%\centering
%\vspace{-15pt}
\hspace{-10pt}
\includegraphics[trim=0cm 0cm 0cm 0cm, clip=true, totalheight=0.26\textheight, angle=0 ] {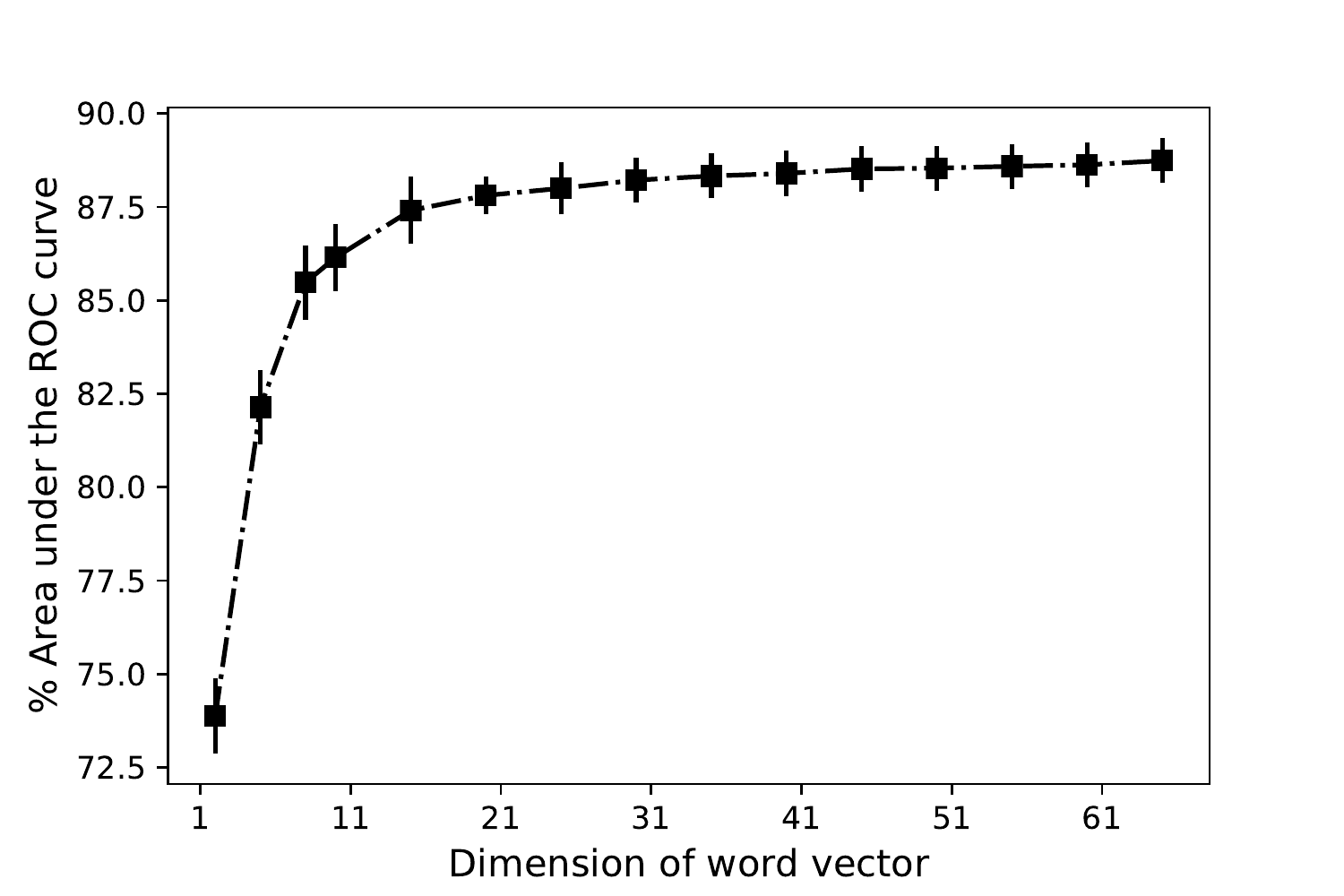}
\hspace{-10pt}
\includegraphics[trim=0cm 0cm 0cm 0cm, clip=true, totalheight=0.26\textheight, angle=0 ] {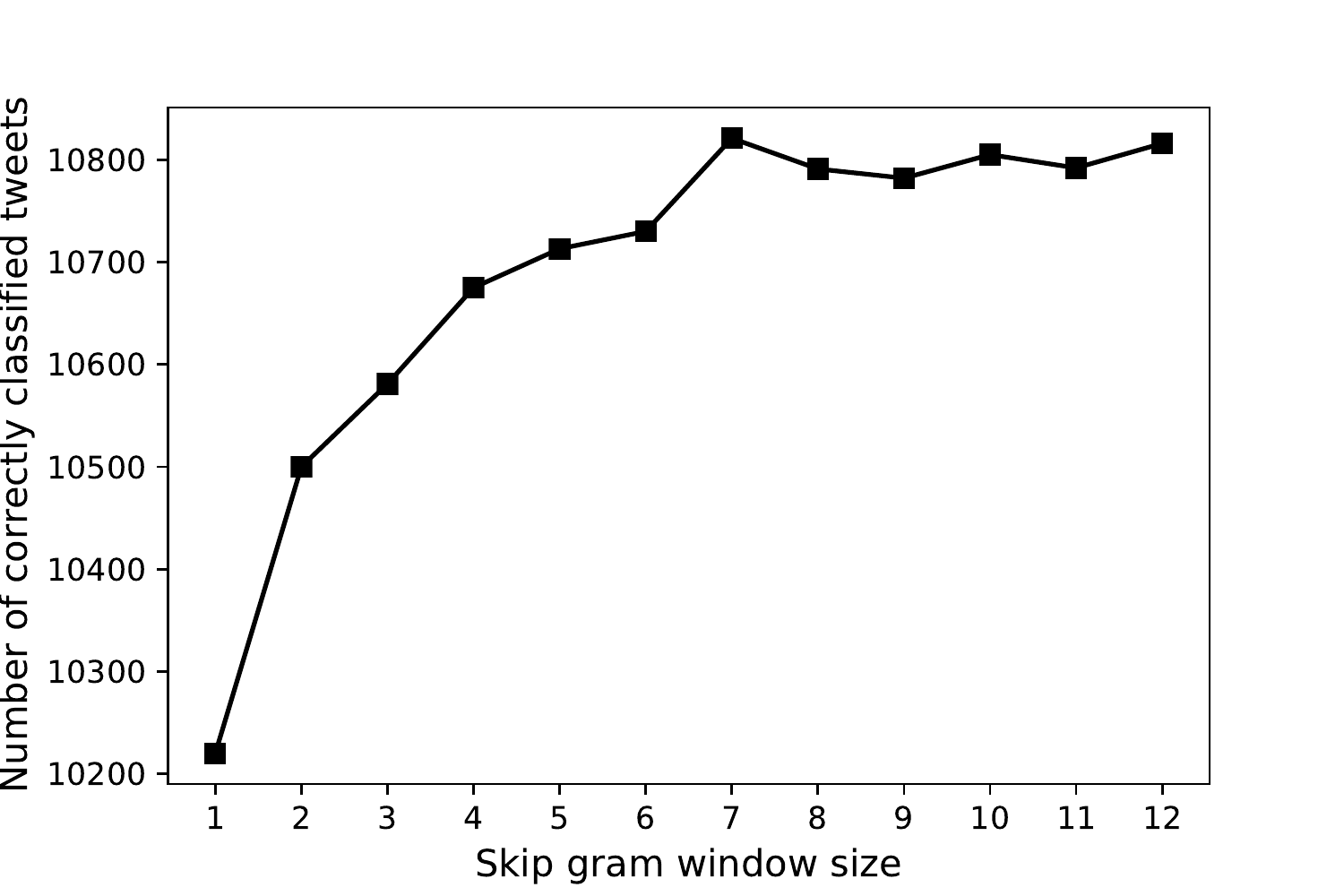}
\caption{Effect of Skip gram parameters: (a) word vector dimensionality and (b) n-gram window size on Twitter sentiment classification performance.}
%\vspace{-20pt}
\label{fig:02}
\end{figure*}

\section {Results}

The Airline Sentiment Twitter data set includes 3,099, 2,363, and 9,178 tweets labeled as neutral, positive, and negative tweets, respectively. The label imbalance in the data set can bias the 3-way classification performance due to the higher proportion of tweets with negative sentiment. With an imbalanced data set, the accuracy metric appears inflated and misleading as the classifier can identify most neutral tweets with negative sentiment. Both our study and~\citep{Jeni2013} suggest similar observation on F1-score and the receiver operating characteristic curve (ROC) as follows.  F1-score and its constituent precision-recall values are significantly affected by data imbalance. The ROC includes sensitivity and specificity values for varying thresholds on classification scores and is not affected by data imbalance. An F1-score or a confusion matrix can be obtained for each point on the ROC. Thus, the ROC captures the trade-off between sensitivity and specificity in area under the curve (AUC) scores. Hence, we use AUC scores to report and compare the classification performances. To train an unbiased classifier model, the training data set (90\%) is adjusted for imbalance using the SMOTE algorithm~\citep{Chawla2002}, which is equivalent to setting class\_weight = 'balanced' in scikit learn classifier modules. However, the left-out test data fold (10\%) is unadjusted for imbalance. Adjusting test samples for imbalance will inflate the test accuracy due to the presence of additional representative class samples from oversampling. For all classification tasks, we have used the logistic regression classifier model with  hyperparameters : the c value (inverse of regularization weight) and regularization type (L1 and L2 norms). Note that more sophisticated nonlinear classifier models, such as random forest or gradient boosting trees, do not improve the performance over the linear classifier model at the cost of higher computational steps as discussed later.   We discuss our findings in the following sections.

\begin{figure}[t]
\normalsize
\centering
%\vspace{-15pt}
\hspace{-15pt}
\includegraphics[trim=0cm 0cm 0cm 0cm, clip=true, totalheight=0.3\textheight, angle=0 ] {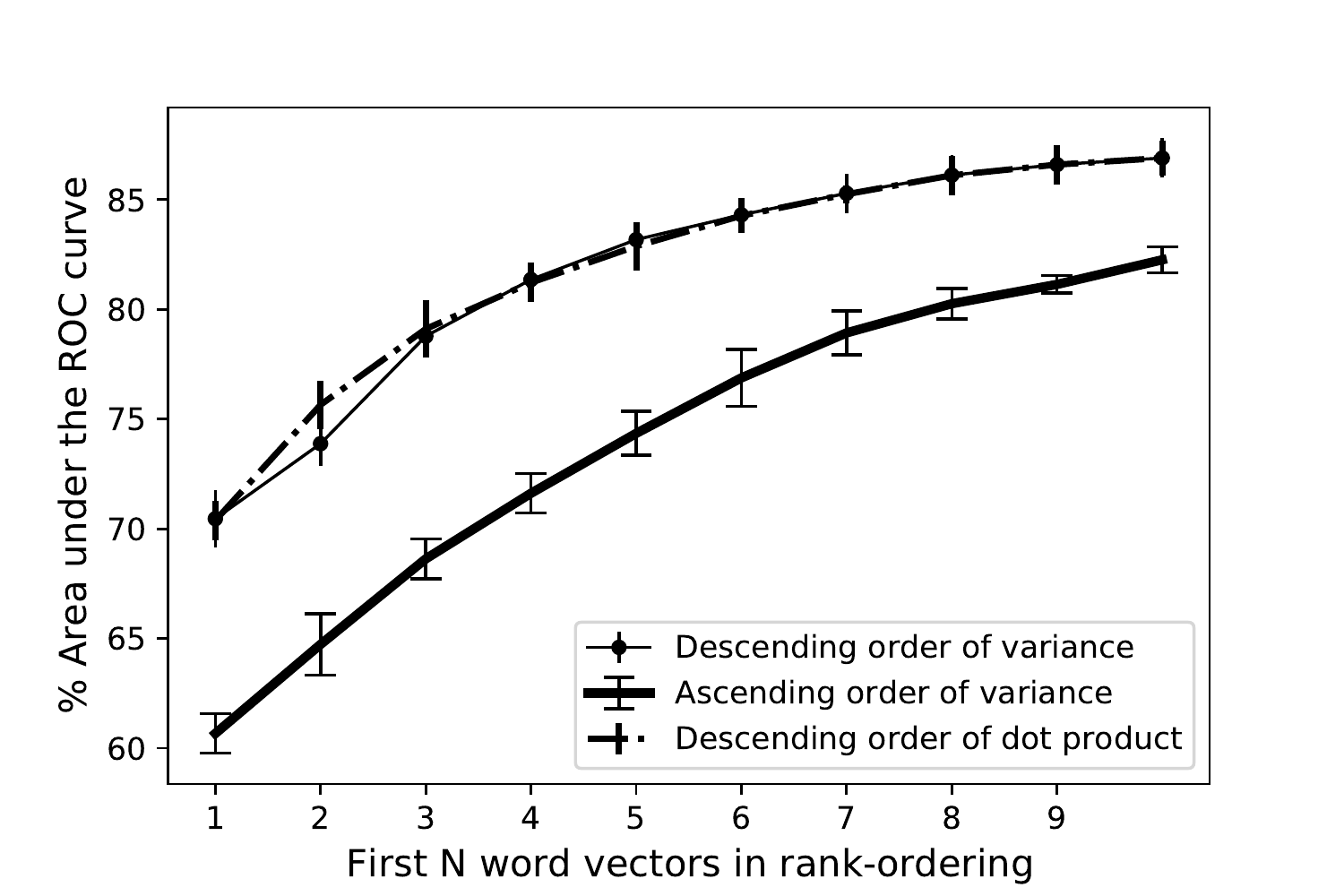}
\vspace{-10pt}
\caption{Effect of selecting N most important words on sentiment analysis. Cumulative contribution of top N words selected from descending or ascending rank order of variance and L2-norm of word vectors. Superior performance of descending order suggests higher variance and dot products infer more informative words.}
%\vspace{-15pt}
\label{fig:03}
\end{figure}

\subsection {Effect of text processing steps}

We identify eight text processing cases (Table~\ref{tab:03}) using four text processing steps that are shown in Table~\ref{tab:02}.

\begin{table*}[t]  
%\centering. 
\caption{Example results of eight Tweet processing cases. Stop word and Signal char denote stop word and single character removal, respectively. The '\#' character is removed keeping the keyword to retain the context of hashtag. Case 3 appears to be the cleanest with most informative tokens.}
\begin{tabular} {lll}
			\toprule
Number & Definition &	Tokenized list of strings\\ \midrule
\multirow{3}{*}{Case 0}   &	No processing  &	['\&amp', '\#gofundme', '\verb+@+united', 'the', 'refund', 'request?', \\ 
&(raw tweet)& "functionality's", 'w', 'on',   'http://t.co/5IMDckODfx', 'is', \\ 
&&'broken.i', 'get', 'a', 'time', 'out', 'error.', \dSmiley] \\ \midrule
\multirow{2}{*}{Case 1}  &	\multirow{ 2}{*}{RegEx} &	['gofundme', 'the', 'refund', 'request', '?', 'functionality', 's', 'w', 'on', \\
&&      'is', 'broken',   'i', 'get', 'a', 'time', 'out', 'error', \dSmiley ] \\ \midrule
Case 2 &	RegEx + Stop word&	['gofundme', 'refund', 'request', '?', 'functionality','s', 'w', 'broken', \\ 
&&'i', 'get', 'a', 'time', 'error', '\dSmiley'] \\\midrule
\multirow{ 2}{*}{Case 3}	& RegEx + Stop word  & ['gofundme', 'refund', 'request',  'functionality', 'broken', 'get', \\ 
 &+ Single Char & 'time', 'error']\\\midrule
\multirow{2}{*}{Case 4}	& RegEx+Single Char&	['gofundme', 'the', 'refund', 'request', 'functionality', 'on', 'is',  \\ 
& + Encode Emoji&'broken', 'get',  'time','out', 'error', b' \verb+\\+xf0\verb+\\+x9f\verb+\\+x98\verb+\\+x8a' ] \\ \midrule
\multirow{2}{*}{Case 5} &	RegEx + Stop word + Single  &	['gofundme', 'refund', 'request', 'functionality', 'broken','get', '\\
& Char + Encode Emoji & time', 'error', b'\verb+\\+xf0\verb+\\+x9f\verb+\\+x98\verb+\\+x8a'] \\ \midrule
\multirow{2}{*}{Case 6} & \multirow{2}{*}{RegEx + Encode Emoji} &	['gofundme', 'the', 'refund', 'request', '?', 'functionality', 's', 'w', 'on', 'is',\\
&&  'broken', 'i', 'get', 'a', time', 'out', 'error', "b'\verb+\\+xf0\verb+\\+x9f\verb+\\+x98\verb+\\+x8a'"]\\ \midrule
\multirow{2}{*}{Case 7} &	RegEx + Stop word  &	['gofundme', 'refund', 'request', '?', 'functionality', 's', 'w', 'broken', \\
&+ Encode Emoji&'i', 'get', 'a', 'time', 'error', "b'\verb+\\+xf0\verb+\\+x9f\verb+\\+x98\verb+\\+x8a'"]\\ \midrule
%&+ Encode Emoji& \\ 
		\end{tabular}
	\label{tab:03}
\end{table*}

Case 0 generates the word-embedding and performs the classification without involving any text processing, which equivalent to simple tokenization of the raw tweet. Case 1 cleans raw tweets using regular expression (regEx) based matching patterns as shown in Table~\ref{tab:01}. This RegEx-based cleaning isolates punctuation and punctuation-spliced words such as 'broken.i' while retaining hashtag words and single characters including the Unicode emoji character showing a happy face. Case 2 additionally removes stop words like ‘on’, ‘out’, and ‘the’. Case 3 builds on Case 2 by additionally removing single characters such as 'a', 'i', 'w', 's' including Unicode characters and emojis. Case 4 includes the stop words (excluded in Case 3) and additionally encodes the Unicode emoji character into its UTF-8 format ("b'\verb+\\+xf0\verb+\\+x9f\verb+\\+x98\verb+\\+x8a'"). Case 5 combines the Case 4 processing steps with stop word removal. Case 6 pairs the procedure from Case 1 with emoji encoding. Case 7 follows the procedure of Case 2 and additionally encodes the emoji character.

\subsection {Word importance and effect of word embedding}
The general intuition is that increasing data dimensionality deteriorates classifier performance due to the 'curse-of-dimensionality' effect. We study the effect of word vector dimensionality on classification accuracy. Figure~\ref{fig:02}(a) shows that higher word vector dimensionality gradually adds more predictive information until it reaches a plateau. There appears to be no significant improvement following the word vector dimensionality of 40. Hence, we report our subsequent analysis results using 40-dimensional word vector. In skip gram model, the number of neighboring words to be predicted from the input word (discussed with Eq. 2) also affects the performance of word embedding. Figure~\ref{fig:02}(b) shows the effect of increasing the window size of neighboring words on Twitter sentiment classification performance. A window size of seven appears to be the best choice for our particular Twitter classification task. 
%%%%%%%%%
The importance of individual words is calculated from the variance and L2 norm of the word vectors. The words with the highest and lowest variance of word vectors yield mean AUCs of 70 and 60, respectively. This suggests that word vectors with higher variance are more informative than those with low variance. The cumulative performance gains after including the top N most important words and the N least important words are shown in Figure~\ref{fig:03}. The top 10 most important words in a tweet yield 98.33\% of the maximum sentiment classification accuracy that is obtained using all available words. The L2 norm and variance of word vectors result in similar accuracy trends as shown in Figure~\ref{fig:03}. 
%%%%%%
\subsection {Effect of tweet embedding}
We compare the performance of three proposed tweet embeddings with that of the baseline vector averaging method in Twitter sentiment classification. Table~\ref{tab:04} shows 10$\times$2 nested cross-validation AUC-ROC for four tweet embeddings and eight tweet processing cases. The baseline averaging of word vectors performs the worst among all tweet embeddings. Case 0, without any text processing, yields low accuracy with the highest standard deviation. The sum of the word vectors yields the most accurate classification of sentiment. However, weighted average slightly outperforms the summation method in several cases and is a superior choice over baseline averaging.  The best classification performance is obtained using text processing Case 6, which is regular expression-based processing followed by encoding of emojis. This suggests the value of encoding and incorporating emojis in Twitter SA. The worst classification performance is obtained using averaging of word vectors in Case 3. Case 3 excludes all the stop words, single characters, and emojis to yield cleanest tokens in Table~\ref{tab:03}. Case 3 text processing scheme invariably yields the worst performance for all types of tweet embedding. This observation suggests that aggressive removal of stop words and characters has negative effects on word embedding, classifier model training and performance. Using the 10 most important words (based on word vector variance) and summation of these word vectors, the classification performance is comparable to other tweet embedding methods.   
%%%%%%%%%
\begin{table}[t]  
\centering. 
\vspace{10pt}
\caption{Effects of eight text processing cases and four tweet embeddings on sentiment classification performance. Values represent average area under the receiver operating characteristic curve (AUC) across 10 cross-validation folds with standard deviation in the parenthesis. AVG = average, WAVG = weighted average, VECSEL = selecting top 10 words from the descending order of word vector variance. The best and worst AUC scores are highlighted for comparison.}
\begin{tabular} {l|llll|c}
			\toprule
Case & 	SUM	&AVG	&WAVG	&VECSEL	& Best \\ \midrule
%&&&&&Embedding\\ \midrule
0&	87.71(1.0)&	86.21(1.2)	&87.47(1.1)&	86.80(1.3)&	SUM\\
1	&88.21(0.5)	&86.48(0.4) & 88.13(0.4) 	&86.35(0.6)	&SUM \\
2	&88.15(0.4)	&86.39(0.4)	&88.18(0.3)&	86.81(0.6)	&WAVG \\
3	&87.31(0.4)	&\bf{85.32(0.4)}	&87.32(0.4)&	86.32(0.6)	&WAVG\\
4	&87.44(0.5)&	85.54(0.6)	      &87.40(0.6)&	86.46(0.8)&	SUM\\
5	&87.52(0.4)&	85.51(0.5)	&87.37(0.4)&	86.49(0.7)&	SUM\\
6	&\bf{88.37(0.4)}&	86.66(0.4) &	88.33(0.4)	 &86.88(0.5)	&SUM\\
7	&88.21(0.5)&	86.65(0.5)	 & 88.28(0.4)	&86.98(0.5)	&WAVG\\ \midrule
Best  	 & Case 6	 &   Case 6  	&   Case 6	  & Case 7	& \\
%case &&&&&\\
\midrule
		\end{tabular}
	\label{tab:04}
\end{table}
%%%%%%%%%
We further compare classification performance for individual sentiment classes. For the best (sum + case 5) and worst (average + case 3) scenarios, we show their corresponding confusion matrices in Table~\ref{tab:05}. The average accuracies of the best and worst scenarios are 73.33\% and 69\%, respectively. The confusion matrices show that negative tweets are confused as positive tweets much less than the positive tweets are confused as negatives. The worst model heavily confuses neutral tweets as negative tweets. Classifying neutral tweets is the most difficult task as it is a gray zone between positive and negative sentiment classes. Hence, the accuracy of neutral tweet classification accounts for most deviations between the best scenario (71\%) and the worst scenario (60\%).

\begin{table}[t]  
\centering. 
\caption{Confusion matrices and F1-scores for three sentiment classification. The best (sum + case 6) and worst (avg + case 3) models (Table~\ref{tab:03}) are used as examples. }
\begin{tabular} {c|cccc|c}
			\toprule
\multirow{ 5}{*}{Best}& Actual & \multicolumn{3}{c|}{Predicted As}& F1-score  \\ \midrule
        &	&	Negative&	Neutral&	Positive&\\
	&Negative	&76\%&	17\%	&7\%& 0.82\\
  Model &Neutral	 &17\% &	71\%	&12\%& 0.60\\
	&Positive	&11\%	&16\%	&73\%&0.67\\ 
	\midrule
	\midrule
\multirow{ 5}{*}{Worst}& Actual& \multicolumn{3}{c|}{Predicted As}& F1-score	 \\ \midrule
 &	&	Negative	&Neutral&	Positive&\\
	&Negative&	78\%	&16\%&	6\%&0.83\\
 Model	&Neutral&	26\%&	60\%&	14\%&0.56\\
	&Positive	&14\%&	17\%&	69\%&0.66\\
\midrule
		\end{tabular}
		\vspace{-5pt}
	\label{tab:05}
\end{table}

\begin{table*}[t]  
\centering. 
\caption{Comparison of classifier model performances in Twitter sentiment classification. Time is in minutes required for the nested cross-validation that includes generation of the word embedding at each fold iteration. SVM-RBF stands for support vector machine with radial basis function kernel.}
\begin{tabular} {llllll}
			\toprule
	& SVM-Linear	&Logistic &	SVM-RBF	&Random&	Gradient Boosting Trees\\
	 &&Regression&&Forest\\\midrule
Hyper & C = [0.01, 0.5, & ['L1', 'L2'] &C = [0.01, 0.5, 0.05, & Num. of trees  &	Num. of trees=[50, 80, 110] \\
para- & 0.05, 0.8, 1, 5] &C = [0.01, 0.5, &$ 0.8, 1, 5],~\gamma$ = [0.1, 0.3,  &=[10, 30, 50, &learning rate = [0.1, 0.5], \\
meter&&0.05, 0.8, 1, 5]&0.5, 0.01, 0.02, 0.05]&70, 90, 110] &max\_depth = [2, 5, 10]\\\\
Time &	70 minutes &	6 minutes	& 756 minutes	& 13 minutes	& 312 minutes\\\\
 &\multirow{3}{*}{88.4 (0.8)} &\multirow{3}{*}{	88.4 (0.4)}	&\multirow{3}{*}{88.6 (0.6)}	&\multirow{3}{*}{87.7 (0.5)}	&\multirow{3}{*}{87.7 (0.6)}\\
AUC 	 &&&&&\\
&&&&&\\
\midrule
		\end{tabular}
	\label{tab:06}
\end{table*}

\begin{table*}[t]  
\centering. 
\caption{ An example of tokenized tweet and ordered list of words based on importance score derived from word-embedding. The trivial words appear with low importance scores. }
\begin{tabular} {ll}
			\toprule	
Tokenized 
Tweet &	['i', 'am', 'appalled', 'at', 'your', 'lack', 'of', 'communication', 'during', 'our', 'awful', \\
& 'journey', 'on', 'ewr', 'to', 'bos', 'hr', 'delays', 'need', 'to', 'be', 'explained', 'better']\\\midrule
Ordered list of words& [journey (0.49), appalled (0.46), explained (0.42), communication (0.31), lack (0.31), \\
with importance scores & BOS (0.24),     awful (0.24),  hr (0.24),	EWR (0.21),during (0.20), \\
 & delays (0.18), better (0.16), our (0.16), need (0.13),am (0.13), be (0.09), at (0.09), \\
&of (0.08), your (0.07), I (0.07), on (0.05), to (0.05)]\\
\midrule
		\end{tabular}
	\label{tab:07}
\end{table*}

\section {Discussion}

This paper compares the performance of different text processing strategies to investigate an optimal pipeline for Twitter sentiment classification. The key findings of this paper are: 1) Reg-Ex based text cleaning is sufficient since removal of stop words and characters negatively affects model training and classification performance, 2) retaining and encoding non-ASCII emoji and previously overlooked characters are important to improve Twitter sentiment classification, 3) sum of word vectors is preferred over averaging to yield tweet vectors for classification, 4) the skip gram model-derived word embedding provides with the information about word importance for data-driven and domain-specific selection of words. In the following subsections, we provide further discussion of our observed results. 

\subsubsection*{Off-the-shelf text cleaning routines affect sentiment classification  performance:} Our results show that skip gram model-derived word embeddings are rich in predictive information and robust to noise in several ways. Even the stop words that are routinely removed in text analysis show low but useful additive predictive information. This is evident in the performance of text processing Case 3. In Table~\ref{tab:02}, Case 3 removes all stop words and single characters to reduce the original tokenized list of 18 tokens (Case 0) to presumably 8 most informative tokens. In fact, Case 3 with the least number of tokens performs the worst among all the cases. Therefore, a basic regular expression-based text cleaning is sufficient (Case 1) without targeting stop words and special characters in character-limited tweet classification. Leveraging special characters like emoji is the key to improving the Twitter sentiment classification performance. This explains why Case 6 (Case 1 plus emoji encoding) appears as the best performing case. On the other hand, averaging the word vectors lowers the variance in the resultant tweet vector. The reduction in data variance may negatively affect sentiment classification performance. This effect is improved by using the weighted average of word vectors where the variance estimate is considered in the average calculation. Overall, the sum of word vectors retains the most predictive information within a tweet vector. This is particularly useful when no vector is available in the word embedding for a test keyword.

The dimensionality of word vectors is often set default as large as 200~\citep{Jianqiang2017} without validating if such large vector dimensionality is the optimal choice. We show that higher vector dimensionality gradually adds more predictive information up to 40 dimensions beyond which computational cost is increased with no added value to the model performance (Figure~\ref{fig:02}(a)). Note that the performance difference in Table~\ref{tab:04} between the cases may appear low in terms of percentage. This is true for the VECSEL and average (avg) based tweet embedding. However, for sum and weighted average (wavg) based tweet vector representations, the standard deviation appears low (0.4\%) with difference in AUC between 1\% and 3\%. This suggests a significant improvement in classification performance across the text processing cases (Case 0 to Case 7). For example, the accuracy difference between the best and worst models (Table~\ref{tab:05}) is 4.33\%. This implies that the best model is able to correctly classify 4.33\% of total 14,640 tweets or 634 tweets more than those of the worst model. This improvement is substantial in terms of sample size and important for real-world application.

\subsubsection*{Linear classifier models are preferable:} The skip gram model obtains quite informative word embedding  that efficiently captures domain contexts. The aggregation (sum or averaging) of these word vectors results in tweet embeddings that yield linearly separable sentiment classes for which a linear classifier model is sufficient.  Unlike linear models, nonlinear classifier models are not only demanding in time and computational resources but also prone to overfitting for this particular application. This is evident in the comparative results obtained using several well known  classifier models in Table~\ref{tab:06}. Both linear models (SVM-Linear and Logistic Regression) yield the same accuracy and slightly outperform the random forest and gradient boosting tree models. The  nonlinear SVM-RBF yields slight improvement at the cost of very high computational time. 

\subsubsection*{Word-embedding for word selection:} The skip gram model derived word vectors (word embedding) contain useful information that can be leveraged to measure individual word importance in a tweet. Since the word embedding is obtained using domain-specific text instead of a global corpus, the word vectors are expected to learn the domain contexts and importance. Individual words in tweets cumulatively add predictive information to the sentiment classification performance as observed in Figure~\ref{fig:03}. Figure~\ref{fig:03} suggests that every word bears a contribution (that may be more or less informative) to the overall classification performance. Therefore, using a default dictionary of stop words for all applications, without recognizing contextual importance of individual words, can limit the overall sentiment classification performance. Instead, we recommend the learning of domain (Twitter) and application (sentiment classification) specific stop words from word embeddings. Table~\ref{tab:07} shows the ordering of all words in a tweet example based on proposed importance scores from word embedding. Intuitively, it shows that  conventional stop words (am, be, at, I, your, on, to) are bearing low importance scores based on their word embedding. Thus, a data-driven approach for stop word selection may better guide for context-specific text classification. Even in the case when some test words are unavailable in the trained word embedding dictionary, existing test words can still cumulatively contribute to the final tweet classification.  
%%%%%%%%
\subsection{Limitations}The Skip gram model, implemented under the word2vec package, is a powerful model for learning the word embedding that ultimately sets the upper limit on accuracy of our Twitter sentiment classification. The goal of this paper is not to propose a new word embedding algorithm or to compare several word embedding algorithms but rather to study the effect and contributions of different text processing schemes when using an efficient word embedding.  Indeed, there are more recent word embeddings that may outperform the skip gram model at the cost of higher computational needs. However, the effects of text processing steps are likely to impact all other word embedding models in more or less a similar fashion that we do not investigate in this paper. The paper compares simple statistics (sum, average, weighted average) to aggregate word vectors into a tweet vector. While this simple strategy works, there are more sophisticated ways to aggregate word vectors into a sentence or tweet vector that are not commonly used in the literature and practice.

Our results cannot be fairly compared with those reported using the same data set for sentiment analysis. A single 80-20 split of samples may yield biased accuracy if the 20\% data include easy-to-classify samples. A mere 10-fold cross validation may not yield separate train-validate-test phases for unbiased model selection and accuracy reporting. Hence, we have proposed 10x2 nested cross-validation to yield an unbiased estimate of classification performance. The overall F1-score, across all three classes, is heavily affected by data imbalance.  The existing literature does not report how the imbalance is adjusted or if the training or entire data set is subjected to imbalance adjustment. The AUC of ROC is unaffected by data imbalance and provides flexibility to choose an appropriate classifier model considering sensitivity and specificity trade-offs as required for a particular application.

\section {Conclusions}

In this paper, we conduct an investigative study of text processing and classification steps to identify optimal strategies for Twitter SA. We conclude that the most aggressive text cleaning routine that removes tokens through stop word and single character removal yields the worst performance. This observation is supported by the finding that all word embedding vectors, including those of stop words, contain additive predictive information. We also find that the inclusion of emoji improves sentiment classification performance. Further analysis finds an optimal dimensionality of word vectors and identifies that the sum of word vectors is superior to averaging for tweet vector formation. These observations recommend a data-driven approach to text processing and word selection as opposed to using off-the-shelf routines and dictionary-based mapping. The skip gram-based word embeddings not only learn domain-specific contexts but also facilitate data-driven informed decision making and alleviate the need for many standard text processing steps. The robustness of skip gram word embedding is evident even in cases when certain words are not in the training corpus. The word embedding also yields a linearly separable feature space that is more suitable for developing fast and generalizable sentiment classifier models.

%\section* {Acknowledgment}
%This work has been partially funded by a grant from Old Dominion University Office of Research Multidisciplinary Funding Program (\#300261-010) and a grant from NSF (ECCS 1310353).  

%\bibliographystyle{unsrt}

\bibliography{twitter}

% that's all folks

\end{document}